\documentclass[letterpaper, 10 pt, conference]{ieeeconf}  % Comment this line out
\IEEEoverridecommandlockouts                              % This command is only
\usepackage{graphics} % for pdf, bitmapped graphics files
\usepackage{epsfig} % for postscript graphics files
\usepackage{amsmath} % assumes amsmath package installed
\usepackage{amssymb}  % assumes amsmath package installed
\usepackage{xcolor}
\usepackage[ruled,vlined]{algorithm2e}
\usepackage{float}
\usepackage{caption}
\usepackage{amsmath}
\DeclareMathOperator*{\argmax}{arg\,max}

\title{\LARGE \bf
Efficient Pig Counting in Crowds with Keypoints Tracking and Spatial-aware Temporal Response Filtering
}

\author{Guang Chen$^{1,2}$, Shiwen Shen$^{1*}$, Longyin Wen$^{1}$, Si Luo$^{1}$ and Liefeng Bo$^{1}$% <-this % stops a space
\thanks{*Corresponding author. {\tt(shiwenshen@engineering.ucla.edu)}}% <-this % stops a space
\thanks{$^{1}$ JD Finance America Corporation, 675 E Middlefield Road, Mountain View, CA 94043, USA}%
\thanks{$^{2}$Department of EECS, University of Missouri, Columbia, MO 65211, USA. Guang Chen contributed to the project during his internship at JD Digits.}%
}

\begin{document}

\maketitle
\thispagestyle{empty}
\pagestyle{empty}

%%%%%%%%%%%%%%%%%%%%%%%%%%%%%%%%%%%%%%%%%%%%%%%%%%%%%%%%%%%%%%%%%%%%%%%%%%%%%%%%
\begin{abstract}
Pig counting is a crucial task for large-scale pig farming, which is usually completed by human visually. But this process is very time-consuming and error-prone. Few studies in literature developed automated pig counting method. Existing methods only focused on pig counting using single image, and its accuracy is challenged by several factors, including pig movements, occlusion and overlapping. Especially, the field of view of a single image is very limited, and could not meet the requirements of pig counting for large pig grouping houses. To that end, we presented a real-time automated pig counting system in crowds using only one monocular fisheye camera with an inspection robot. Our system showed that it produces accurate results surpassing human. Our pipeline began with a novel bottom-up pig detection algorithm to avoid false negatives due to overlapping, occlusion and deformation of pigs. A deep convolution neural network (CNN) is designed to detect keypoints of pig body part and associate the keypoints to identify individual pigs. After that, an efficient on-line tracking method is used to associate pigs across video frames. Finally, a novel spatial-aware temporal response filtering (STRF) method is proposed to predict the counts of pigs, which is effective to suppress false positives caused by pig or camera movements or tracking failures. The whole pipeline has been deployed in an edge computing device, and demonstrated the effectiveness.

% Pig counting is critical in current large-scale agricultural production management and asset estimation, which is useful to improve management in pigs feeding and piggery construction. Traditional farm relies on manual counting to obtain the accurate number of pigs, which is inefficient and challenging due to several factors, such as occlusion, overlapping, and similar appearance. In this paper, we provide a solution to handle the pig counting problem using an automatic inspection robot to scan pigsties.  Specifically, we use the deep convolutional neural network (CNN) to extract predefined keypoints of pigs, followed by the non-maximal suppression post-processing. Then, a greedy search algorithm is used to associate the keypoints of pigs in video sequences. After that, we design a cross-line counting algorithm to produce the accurate number of pigs. Extensive experiments demonstrate the effectiveness of our system, which is an important step towards accurate real-time pig counting.
\end{abstract}

%%%%%%%%%%%%%%%%%%%%%%%%%%%%%%%%%%%%%%%%%%%%%%%%%%%%%%%%%%%%%%%%%%%%%%%%%%%%%%%%
\section{INTRODUCTION}

Frequently counting the number of pigs in grouping houses is a critical management task for large-scale pig farming facilities. On one hand, pigs are often moved into different barns at distinct growth stages or grouped into separate large pens by size. Farmers need to know how many pigs are in each large pens. On the other hand, comparing the counting result with the actual number of pigs enables the early detection of unexpected events, e.g., missing pigs. However, walking around the pig barns to count a large number of pigs is costly in labor. Thus, automated pig counting and monitoring using computer vision techniques is a promising way to support intensive pig farming management, while reducing cost. 

\begin{figure}[t]
	\centering
	\includegraphics[trim=0 380 480 0, clip, width=0.99\linewidth]{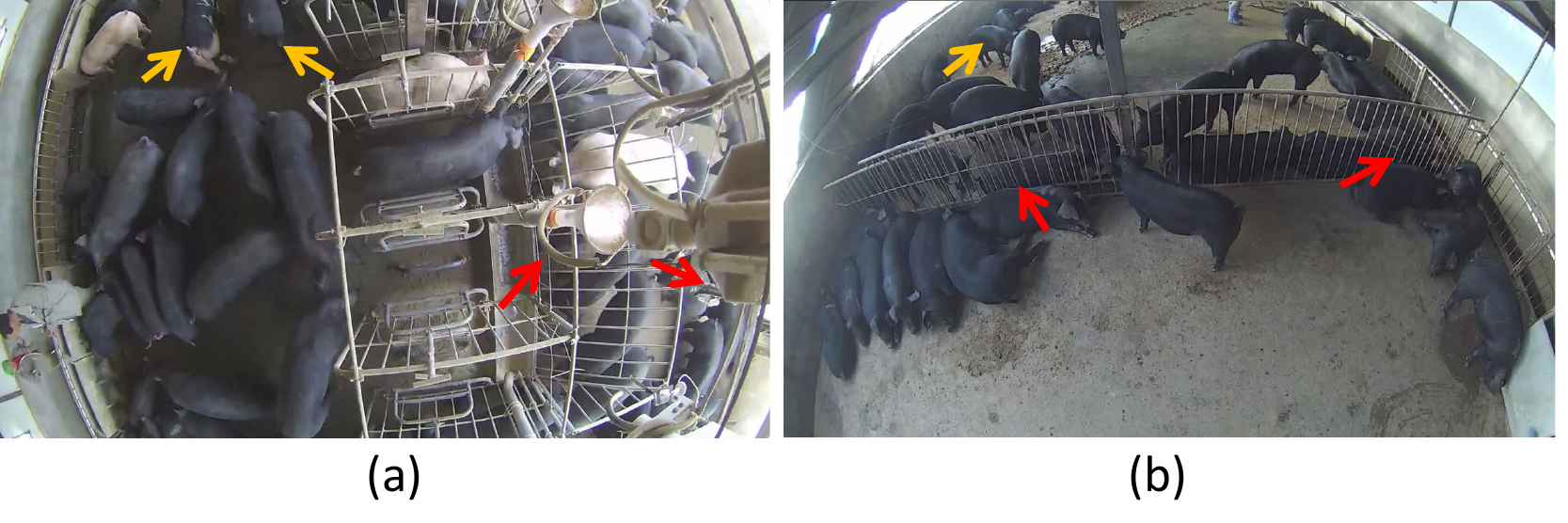} % requires the graphicx package
	\caption{Illustrations of pig counting challenges in large grouping houses. The top-down view images are captured by our inspection robot with a fisheye camerat. Red arrows point to examples of pig overlapping and occlusion. Yellow arrows show cases where pigs are moving in or out of camera field of view.}
	\label{pig_example}
\end{figure}

\begin{figure}[t]
	\centering
	\includegraphics[trim=0 360 480 0, clip, width=0.99\linewidth]{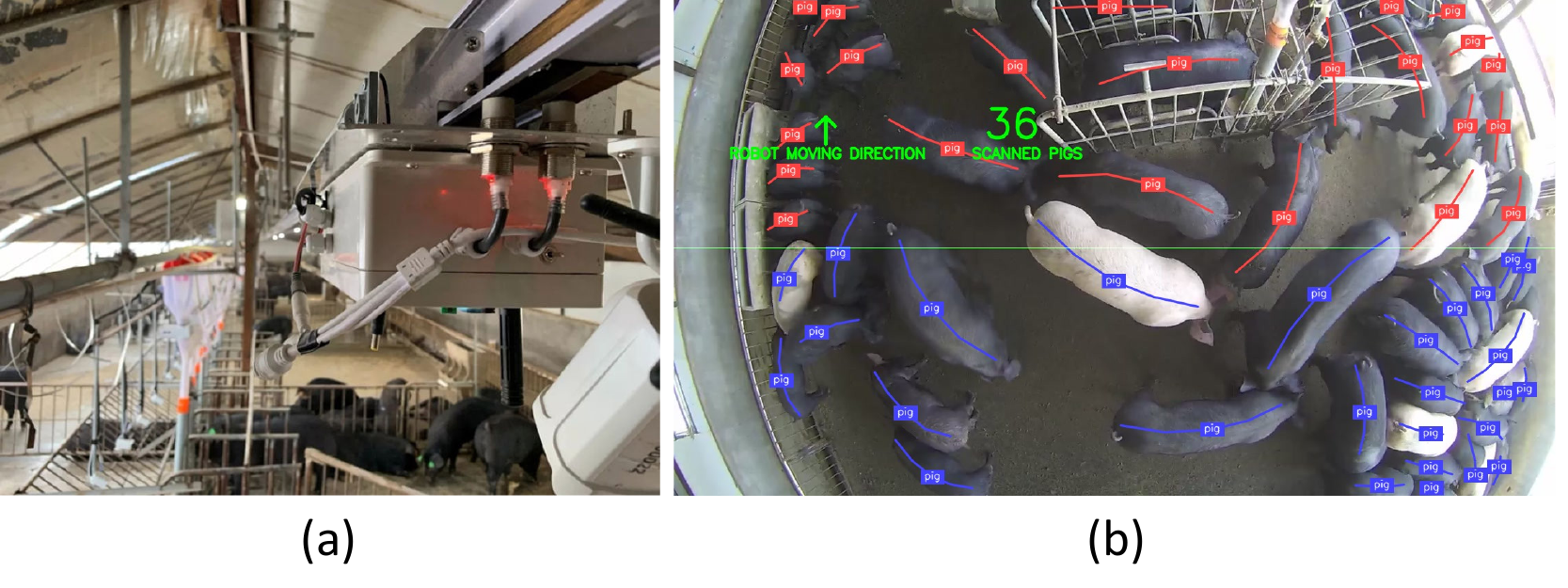} % requires the graphicx package
	\caption{Illustrations of our pig counting system. (a) the installed inspection robots with rails and fisheye cameras for pig counting. (b) a single video frame with detected pig skeletons using our counting algorithm.}
	\label{system_example}
\end{figure}

In recent years, various computer vision algorithms have been widely adopted to support various developments of agriculture and farming automation, such as cattle gait tacking \cite{gardenier2018object}, pig weight estimation \cite{DBLP:journals/cea/PezzuoloGSGM18} and fruit counting \cite{liu2019monocular}. Despite of these exciting progresses, pig counting remains a very challenging task, due to large pig movements, high group density, overlapping, occlusion and camera perspective, as illustrated in Fig.~\ref{pig_example}. Few works in literature studied the development of automated pig counting system. Existing works \cite{DBLP:journals/cea/TianGCWLM19} only handled pig counting problem in a single image. Nonetheless, as shown in Fig.~\ref{pig_example}, the field of view of a single image is only restricted to a small region and it is impossible to monitor a large pig grouping house. Furthermore, it could not deal with the cases that pigs frequently enter into or exist from the camera view. Towards overcoming these challenges, we presented an novel automated counting algorithm with an inspection robot and monocular fisheye camera. Fig.~\ref{system_example}a showed two pictures of our inspection robot with a fisheye camera installed on the roof rail in our experimental pig grouping houses. Fig.~\ref{system_example}b visualized a single video frame with detected pig skeletons output using our pig counting pipeline.

\begin{figure*}[t]
	\centerline{\includegraphics[trim=0 200 0 0, clip, width=1.0\textwidth]{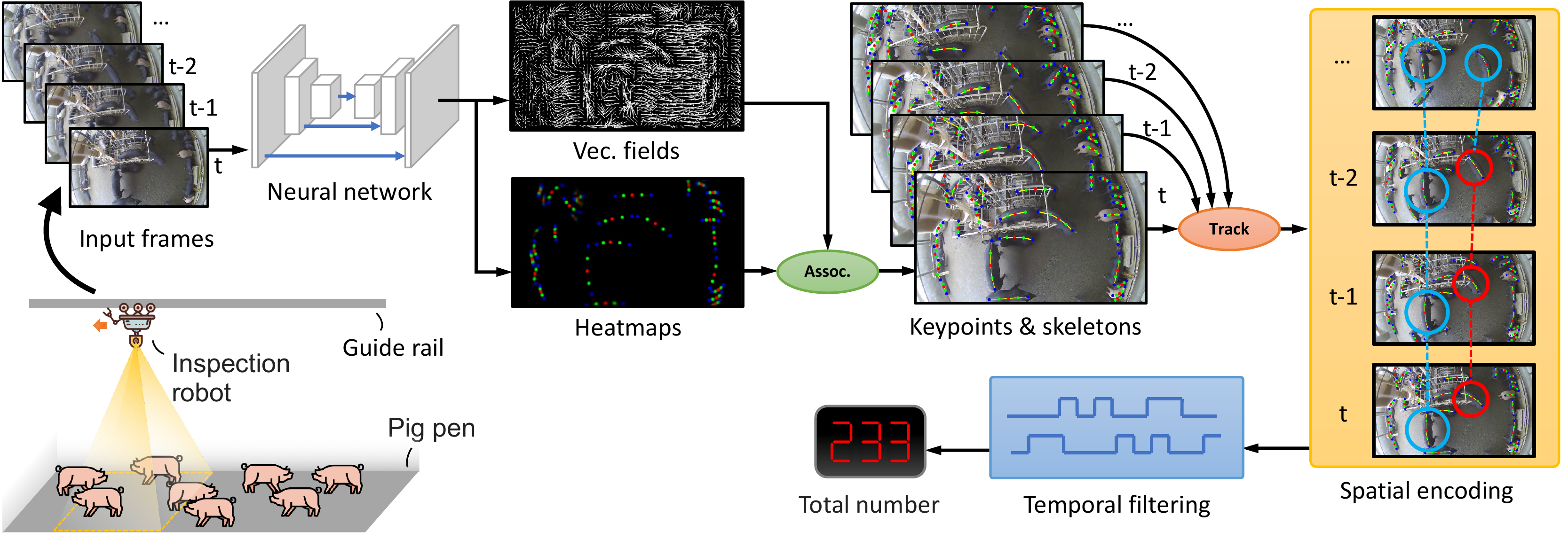}}
	\caption{Pig counting pipeline. The inspection robot moved from one side of the pig house roof to the other end to scan the whole region. A proposed bottom-up detection CNN model was first applied on each video frame to obtain the keypoints and skeletons of all pig candidates. An on-line tracking algorithm was then used to generate the temporal associations across frames. Lastly, STRF, including spatial encoding and temporal filtering, was used to generate the final count.}
	\label{fig:system}
\end{figure*}

The main contributions of this paper are summarized as follows. 1) The sensor configuration is presented, which is suitable for pig counting in large-scale grouping house. 2) A novel bottom-up detection method is proposed to identify pigs, while addressing detection challenges due to overlapping, occlusion and deformation of body shapes. 3) A novel online spatial-aware temporal response filtering (STRF) method is designed to suppress false positives caused by tracking failures or pig movements. 4) An efficient algorithm of the counting pipeline is designed and deployed to an embedded system, which achieves high speed performance.

\section{RELATED WORK}

Counting in the crowd is an important, but challenging task due to severe occlusion, perspective distortions, complex illumination and diverse distribution of target sizes \cite{shen2018crowd}. Recently, deep-learning-based methods \cite{shen2018crowd, sindagi2017generating} have been developed to estimate single image density map for crowd counting. Sindagi et. al. \cite{sindagi2017generating} developed a contextual pyramid convolutional neural network (CNN) for crowd density map estimation. Both global and local contexts were employed in the network to achieve better accuracy. Shen et. al. \cite{shen2018crowd} proposed an adversarial cross-scale consistency pursuit method to improve the estimation consistency and reduce the averaging effect in \cite{sindagi2017generating}. These methods formulate the counting problem as density map estimation, thus having the advantage to handle server occlusion and perspective distortions. However, density-map-based methods lost the detailed individual information and discarded the accurate location information for each single target. Therefore, it loses the ability to associate targets across time, and is not suitable for video-based counting.

Recently, researchers in agriculture presented many works towards tackling counting problems in various scenarios. Tian et. al. \cite{DBLP:journals/cea/TianGCWLM19} counted pigs in a single image using a CNN-based method for pig density map estimation. Similar as \cite{sindagi2017generating}, this method is not suitable for video-based counting problem. As a single image only have a small field of view (as shown in Fig.~\ref{pig_example}), it cannot be used for pig counting in large grouping houses. Liu et. al. \cite{liu2019monocular, liu2018robust} developed a fruit counting pipeline using a monocular camera. Individual fruits are first segmented using a CNN-based method, and then tracked by a Kalman Filter corrected Kanade-Lucas-Tomasi (KLT) tracker. A structure from motion (SfM) algorithm was utilized to get the relative 3D location and size estimate to reject outliers and double counted fruit tracks. This method is only suitable for rigid shape and stationary target counting task, and does not work for moving livestock counting cases. Hodgson et. al. \cite{hodgson2016precision} demonstrated that images collected by unmanned aerial vehicles (UAV) could help wildlife monitoring and counting. Rivas et. al. \cite{rivas2018detection} studied cattle detection from aerial view photos. Counting based on aerial view is promising, but could hardly be used for indoor livestock counting scenes without developing algorithms to handle severe occlusion (e.g. caused by indoor building structures or perspective distortions), overlapping, double counted tracking trajectories due to entering into or existing from camera view.

Different from previous approaches, we presented a novel video-based pig counting system for large pig grouping houses. The developed counting pipeline overcame dense detection challenges (e.g. overlapping or occlusion) by a novel bottom-up pig body parts detection and association algorithm. A STRF method was developed to obtain the counting number by reducing the counting error caused by tracking failures or pig movements.

\section{APPROACH}

In this work, we presented an efficient pig counting system for large pig grouping houses. Fig.~\ref{fig:system} demonstrates the entire algorithm pipeline. In our counting system, the camera moved from one side of the pig grouping house roof till the other end of roof and scanned the whole house with top-down view. A whole single counting pass scanned the house once by the camera. As summarized in the Fig.~\ref{fig:system}, subsequently, we detected multiple pig body keypoints, associated them to localize each individual pig, tracked pigs cross frames and obtained counting results using STRF method.

\subsection{SENSOR CONFIGURATION}

An inspection robot, which can move back and forth along a rail installed on the roof of the pig house, was used for pig counting data acquisition and processing from top-down view (Fig.~\ref{system_example}a). Several sensors used for different applications were installed inside the robot, including a monocular fisheye camera for pig counting, a RGB-D camera for pig weight estimation, gas and temperature sensors for environmental control etc. Inside the inspection robot, an embedded system with RockChip RK3399 multi-core ARM processor was used for processing data from cameras and running the pig counting algorithm. 

\subsection{BOTTOM-UP DETECTION}
\begin{figure}[t]
	\centerline{\includegraphics[trim=0 230 490 0, clip, width=0.49\textwidth]{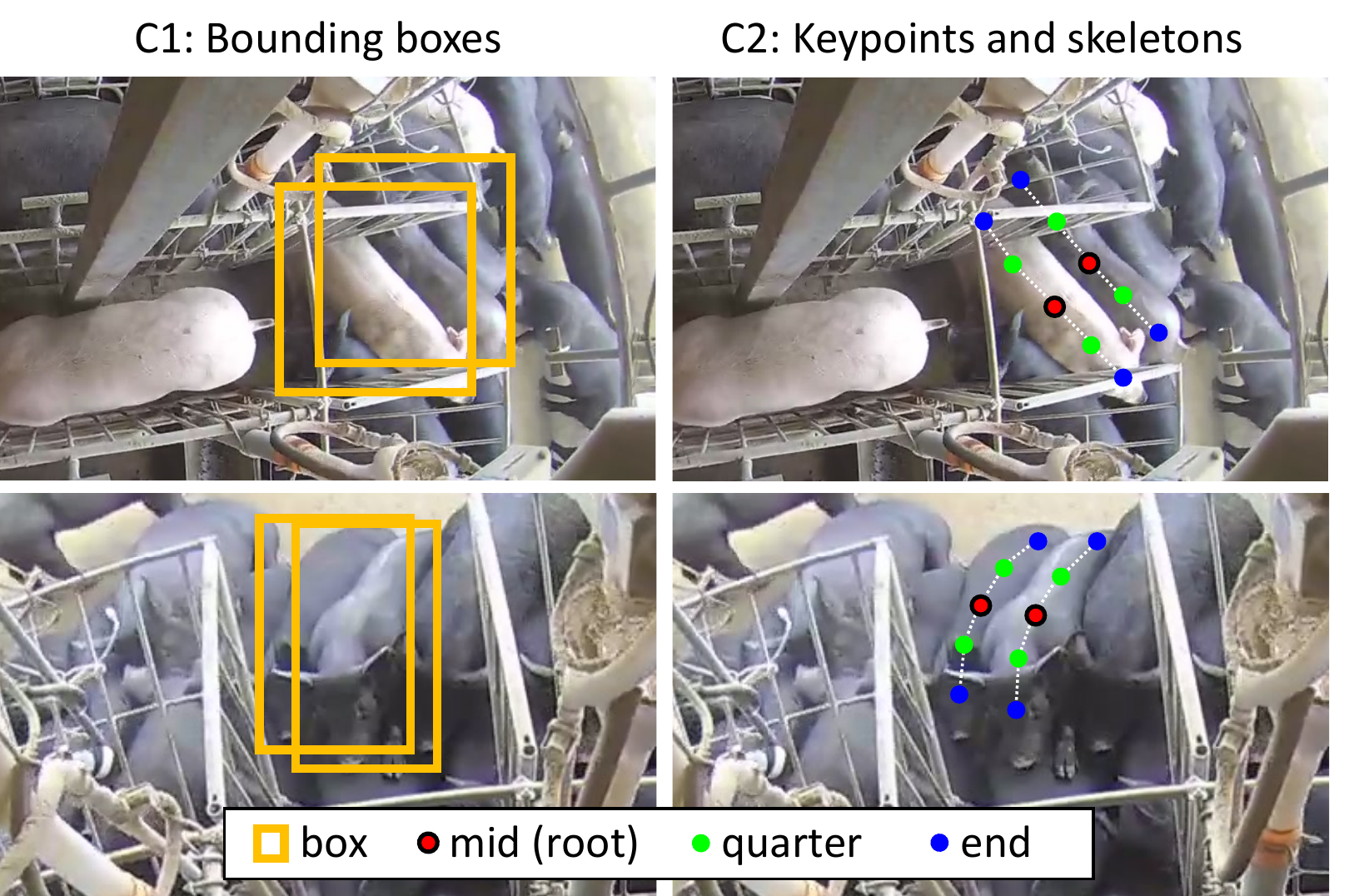}}
	\caption{Illustration of top-down bounding boxes v.s. bottom-up keypoints for pigs detection. Column 1 (C1): bounding boxes had very high overlap ratios for adjacent pigs. Column 2 (C2): body parts keypoints for adjacent pigs. In this work, five keypoints are defined: one middle body part keypoint, two body end keypoints and two quarter body keypoints.}
	\label{fig:box_sucks}
\end{figure}

\begin{figure*}[t]
	\centerline{\includegraphics[trim=0 380 0 0, clip, width=0.95\textwidth]{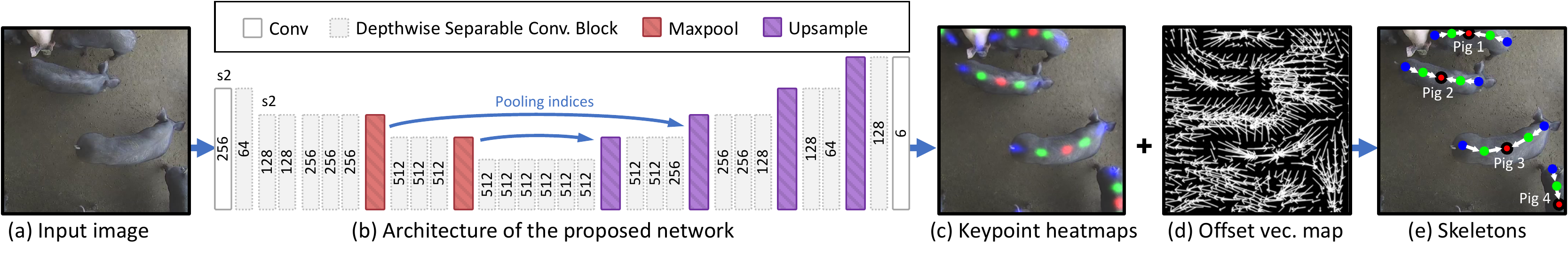}}
	\caption{The proposed bottom-up keypoints detection CNN architecture. The network contains two regular convolutional layer at the two ends and 24 depthwise separeble conv. blocks in-between. In (b) the numbers indicate the number of output channels; And $s2$ means a stride of 2. All the conv. layers except the pointwise conv layers have kernel size 3 $\times$ 3.
	%The numbers in (b) indicate the number of feature maps. $s2$ means the stride is 2.
	}
	\label{fig:arc}
\end{figure*}

The first step was to detect pig candidates in each video frame. Traditionally, top-down object detectors, such as faster RCNN\cite{ren2015faster}, SSD~\cite{liu2016ssd} and YOLOv3~\cite{redmon2018yolov3}, have been widely used. These methods first proposed locations of detection candidates using bounding boxes, and then classified each box to be the real target or not. Non-maximum suppression (NMS) are employed as a post-processing method to significantly reduce false positive candidates by removing the bounding boxes that have high overlap ratios (intersection over union) with each other. Nonetheless, using bounding boxes to localize the pigs is sub-optimal in this application. The deformable long oval pig shapes are very challenging for bounding-box-based approaches in crowded scene. As shown in Fig.~\ref{fig:box_sucks}C1, the bounding boxes around two adjacent pigs have very high overlap ratio, whose ambiguous nature tends to confuse the neural network training. Moreover for inference, the NMS post-processing step would enforce the detector to only select one bounding box for these high overlapping cases, resulting in false negatives. Compared with bounding boxes, the pig skeletons defined by keypoints are more suitable for differentiating pigs in the crowd as shown in Fig.~\ref{fig:box_sucks}C2. In this work, we defined five pig body keypoints, including one \textit{mid point} (red), two \textit{quarter points} (green) and two \textit{end points} (blue); and tree-structured pig skeletons connecting the adjacent keypoints.

% \begin{figure}[htbp]
% 	\centerline{\includegraphics[trim=0 370 700 0, clip, width=0.2\textwidth]{figures/linestrip.pdf}}
% 	\caption{Keypoints on a linestrip along the spine of a pig}
% 	\label{fig:linestrip}
% \end{figure}

% \begin{figure}[t]
% 	\centerline{\includegraphics[trim=0 32 460 0, clip, width=0.5\textwidth]{figures/trajectory.pdf}}
% 	\caption{Illustration of STRF method. (a) spatial encoding defined zones with an example of count $1$ tracking trajectory. The borderline (in green) between the activated zone (in red) and the deactivated zone (in blue) is refered as activity scanning line in the paper. (b) example of count $-1$ tracking trajectory. (c)(d)(e) examples of count $0$ tracking trajectory.}
% 	\label{fig:trajectory}
% \end{figure}

\begin{figure*}[t]
	\centerline{\includegraphics[trim=0 75 0 0, clip, width=0.9\textwidth]{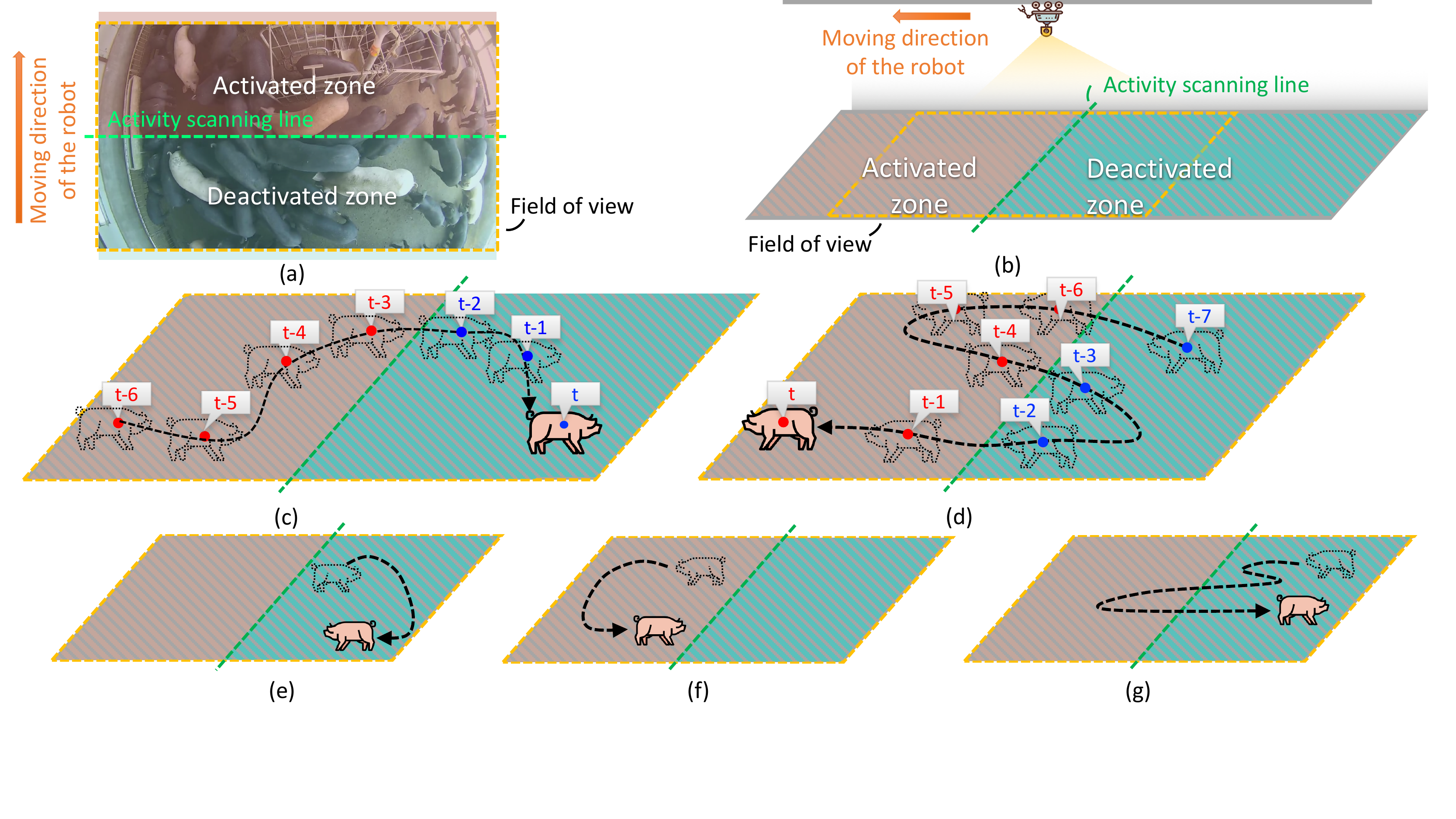}}
	\caption{Illustration of STRF method. (a) spatial encoding defined activated zone and deactivated zone, and an activity scanning line. (b) Activity scanning line moved with the camera to scan the whole pig house. (c) An example of pig tracking trajectory with count $1$. (c) An example of count $-1$ pig tracking trajectory. (d)(e)(f) Examples of count $0$ tracking trajectory.}
	\label{fig:trajectory}
\end{figure*}

Inspired by \cite{cao2017realtime}, we presented an efficient bottom-up detection approach (Fig.~\ref{fig:arc}) to overcome aforementioned limitations. This method is consisted of a keypoints detection step and a keypoints association step. These steps were based on a deep convolutional encoder-decoder network. The network output two different kinds of maps: 1) Four keypoint heatmaps and 2) an offset vector field. (Fig.~\ref{fig:arc}d) Each heatmap provided information to classify each pixel into to one of the keypoints or background class. The offset vector field indicated the relative positional relationships between the adjacent keypoints, which helped the system group the keypoints and identify which pig instance that these keypoints belonged to.

\textbf{Keypoint detection} The goal is to detect all visible keypoints belonging to each single pig in the input. For this purpose, we applied a fully convolutional network to produce heatmaps with four channels (three channels for each keypoint type and one for background), which had the same size with the input image. This heatmap prediction was then formulated as a per-pixel multi-class classification problem. For each pixel location, the neural network learned to predict if it belonged to one of the keypoint type or background. We followed \cite{papandreou2018personlab} to generate classification targets. Let $\mathcal{D}_R(y)~=~\{x:\left\Vert x-y \right\Vert < R\}$ be a circular region centered at position $y$ with radius $R$. We denoted $k_{i,c}$ as the $i$-th keypoint of type $c$. All pixels of $\mathcal{D}_R(k_{i,c})$ had the same class label $c$. In this work, $R$ was set to be 5. Cross-entropy loss was employed for this task. At testing stage, the local maxima of the heatmaps were chosen as the predicted keypoints.

\textbf{Keypoint association}
Due to instance-agnostic nature of the predicted keypoints on heatmaps, one unique instance ID had to be assigned for each detected keypoint so that we "connect the dots" belonging to the same individual instance. For this purpose, we added to our neural network a separate two channel outputs of offset field indicating the displacement from a given keypoint to its parent in the skeleton (Fig.~\ref{fig:box_sucks}C2). Here we denoted $F(k_{i,c})$ as the parent node of keypoint $k_{i,c}$. If $x \in \mathcal{D}_R(k_{i,c})$, the target offset $V(x)$ was vector starting from $x$. If $k_{i,c}$ itself is a root node, i.e. $c = mid$, $V(x)$ ended at $k_{i,c}$; Otherwise $V(x)$ ended at $F(k_{i,c})$.

% when there are multiple pigs in the field of view, extra visual cues are needed to group together the keypoints by the pig they belongs to. Given a predicted keypoint, the system is expected to know where to find its adjacent keypoint on the same pig. We hence design an offset vector field to help with the searching.
% \textcolor{red}{tree-structured pig skeleton}  as well as an \textcolor{red}{tree-structured pig skeleton}. The root node of the tree is the mid point of the pig body. It has two children defined as the two quarter points. A quarter point is the parent node of the neighbour end point. 
% The offset field is a two-channel 2D feature map indicating the displacement from a given keypoint to its parent in the skeleton. (Fig.\textcolor{red}{X})  Here we denote $F(k_{i,c})$ as the parent node of keypoint \textcolor{red}{$k_{i,c}$}. If $x \in \mathcal{D}_R(k_{i,c})$, the target offset

% \begin{equation}
% V(x) = 
% \begin{cases} 
% k_{i,c} & \text{if } c = mid \\
% F(k_{i,c}) - x & \text{otherwise.}\\
% \end{cases} 
% \end{equation}
% $V(x)$ is vector starting from $x$. If $k_{i,c}$ itself is a root node, i.e. $c = mid$, $V(x)$ ends at $k_{i,c}$; Otherwise $V(x)$ ends at $F(k_{i,c})$.

Let us denote the offset field predicted by the network as $U(x)$. In order to supervise the training, the regression loss for offset field was defined as
\begin{equation}
L_r = \sum_{x} (1-G_0(x))\left\Vert U(x)-V(x) \right\Vert^2,
\end{equation}
where $G_0(x)$ was the binary background mask used for ignoring the regression loss at the background pixels, where the offset vector were undefined.

At testing stage, an iterative greedy algorithm was adopt to associate the predicted keypoints. We alternatively searched the best candidate parent node for all the predicted keypoints, and removed the surplus keypoints from their candidate children list, until no better hypothesis could be found. The best candidate parent node was defined as the keypoint which was in the correct class and match the predicted offset vector best. The euclidean distance between the predicted offset and the actual offset was used to measure the match. 

\textbf{Architecture of the network}
We proposed an architecture (Fig.~\ref{fig:arc}b) for the network. Depthwise separable convolutions~\cite{howard2017mobilenets} were used as the basic building blocks to reduce the computational cost. Following~\cite{badrinarayanan2017segnet}, we used location-withheld maxpooling to improve the localization accuracy, which preserved indices at the max pooling layers of the encoder and passed them to the corresponding up-sampling layers of the decoder. 

\begin{table*}[ht]
\centering
\caption{ Comparison of keypoint detection results.}
\begin{tabular}{c|ccc|ccc|c|c}
\hline
Metric    & \multicolumn{3}{c|}{PDJ@0.1} & \multicolumn{3}{c|}{PDJ@0.2} & FLOPs & Parameters \\ \hline
Keypoint type  & mid     & quarter   & end    & mid     & quarter  & end  & &  \\ \hline
%SegNet    & 0.975   & 0.981     & 0.955  & 0.992   & 0.994    & 0.980  & 138G & 29M  \\
UNet \cite{DBLP:conf/miccai/RonnebergerFB15}      & 0.938   & 0.935     & 0.895  & 0.971   & 0.980    & 0.953 & 23G & 42M \\
Hourglass \cite{newell2016stacked} & 0.913   & 0.905     & 0.866  & 0.954   & 0.968    & 0.932 & 23G & 3.6M \\
Ours    & \textbf{0.962}   & \textbf{0.964}     & \textbf{0.934}  & \textbf{0.991}   & \textbf{0.992}    & \textbf{0.978}  & \textbf{15G} & \textbf{3.3M}\\ \hline
\end{tabular}
\label{tab:keypoint_performance}
\end{table*}

\subsection{KEYPOINTS TRACKING}

In order to count pigs across video frames, an efficient on-line tracking method was employed to associate pig keypoints temporally. This method took the grouped pig keypoints for single frames as input, and then assigned a unique identification number (id) to each pig across frames. This problem was formulated as a bipartite graph matching based energy maximization problem. The estimated pig candidates $C^{t}$ at frame $t$ were then associated with the previous pig candidates $C^{t-1}$ at frame $t-1$ by bipartite graph matching. 

\begin{equation}
\begin{split}
\widehat{s} = \argmax_{s}  \sum_{C_{i}^{t} \in C^{t}} \sum_{C_{j}^{t-1} \in C^{t-1}} \Psi_{C_{i}^{t}, C_{j}^{t-1}} \times s_{C_{i}^{t}, C_{j}^{t-1}} \\ 
s.t. \quad \forall C_{j}^{t-1} \in C^{t-1}, \sum_{C_{i}^{t} \in C^{t}} s_{C_{i}^{t}, C_{j}^{t-1}} \in \{ 0, 1 \}, \\
\quad \forall C_{i}^{t} \in C^{t}, \sum_{C_{j}^{t-1} \in C^{t-1}} s_{C_{i}^{t}, C_{j}^{t-1}} \in \{ 0, 1 \},
\end{split}
\end{equation}

\noindent where $C_{j}^{t-1}$ was the $j^{th}$ pig candidate in $C^{t-1}$ and $C_{i}^{t}$ was the $i^{th}$ pig candidate in $C^{t}$. $s_{C_{i}^{t}, C_{j}^{t-1}} \in \{ 0, 1 \}$ was a binary variable and indicates if $C_{j}^{t-1}$ and $C_{i}^{t}$ were associated. The potential $\Psi_{C_{i}^{t}, C_{j}^{t-1}}$ represented the similarity measurements between $C_{j}^{t-1}$ and $C_{i}^{t}$.

\begin{equation}
\Psi_{C_{i}^{t}, C_{j}^{t-1}} = \lambda_{1}\Psi^{A}_{C_{i}^{t}, C_{j}^{t-1}} + \lambda_{2}\Psi^{L}_{C_{i}^{t}, C_{j}^{t-1}},
\end{equation}

\noindent where $\Psi^{A}_{C_{i}^{t}, C_{j}^{t-1}}$ represented the keypoints appearance similarities between candidates. And $\Psi^{L}_{C_{i}^{t}, C_{j}^{t-1}}$ implied the spatial similarities. $\lambda_{1}$ and $\lambda_{2}$ were hyper-parameters to balance the contributions of the two terms.

The spatial similarities was calculated as the $l_2$ distance between the propagated $C_{j}^{t-1}$ spatial location and encoded $C_{i}^{t}$ center location. $\Psi^{L}_{C_{i}^{t}, C_{j}^{t-1}} = \left \|P(L(C_{j}^{t-1})) - L(C_{i}^{t}) \right \|^2 $. The appearance similarity was calculated as the the $l_2$ distance across all keypoints embedded deep features between $C_{i}^{t}$ and $C_{j}^{t-1}$.

\begin{equation}
\Psi^{L}_{C_{i}^{t}, C_{j}^{t-1}} = \sum_{n=1}^{5} \lambda_{n}^{L}\left \|K^{n}_{C_{j}^{t-1}} - K^{n}_{C_{i}^{t}} \right \|^2,
\end{equation}

\noindent where $K^{n}$ represented the $n^{th}$ keypoint deep appearance feature obtained from convolution layer before the last upsampling layer of our keypoints CNN. $\lambda_{n}^{L}$ were the hyper-parameters balancing the weights. 

The aforementioned bipartite graph matching problem was solved using Hungarian method.

\subsection{SPATIAL-AWARE TEMPORAL RESPONSE FILTERING}

Traditionally, video-based counting methods \cite{liu2019monocular, liu2018robust} counted the number of unique tracklet ID as the final counting results. These methods were suitable for the cases, where the target objects were stationary and object occlusion was very rare. In the large-scale pig counting scenario, however, pigs moved fast in different directions, and the same pig will often walked out of the camera view and came back again. In addition, the indoor building structures (e.g. the feeding machine) would sometimes block large part of the camera view causing severe occlusions. Occlusions across long frames will cause tracking failure, and break trajectory of one single object into two or more. In these cases, counting the number of unique tracklet IDs would suffer from large false positive errors. To overcome these limitations, we represented a novel spatial-aware temporal response filtering (STRF) method to perform on-line counting, while minimizing the false positives.

The STRF took the tracking trajectories for all previous frames as input, and output the final counting number. It consisted of two steps: 1) spatial encoding; and 2) temporal response filtering. The spatial encoding stage processed each video frame independently, and each detected pig candidate in the frame was assigned a code number based on their spatial locations. The temporal response filtering stage examined each candidate's trajectory across time and obtained a count number, $count_{i} \in \{ 0, 1, -1 \}$, for this single candidate. The final counting result was the sum of all count number for all candidates: $\sum_{i=0}^{N}count_{i}$.

As shown in Fig.~\ref{fig:trajectory}a, the spatial encoding stage divided one image frame into activated zone and deactivated zone by an activity scanning line. This scanning line was stationary in a single frame, but served to scan the whole pig house moving with the inspection robot. For all detected pig candidates, activity codes will be assigned based on which activity zone these pigs were in. In our work, pigs in activated zone were assigned code value $0$, and pigs in deactivated zone were assigned code value $1$. Deactivated zone indicated that all candidates inside have already been counted by the algorithm; and the candidates in activated zone would be counted when the activity scanning line scanned through them. 

 In the temporal response filtering step, lists of spatial codes in temporal order were generated for each trajectory. One trajectory had one list of spatial codes, and each element of the list corresponded to a time point. Fig.~\ref{fig:trajectory}c illustrated one example of one single pig trajectory from time point $t-6$ to time point $t$, where the blue color represented code $1$ and the red color represented code $0$. As it was shown, the generated temporal code was [0, 0, 0, 0, 1, 1, 1] from $t-6$ to $t$. The final count for this trajectory $count_{i}$ was obtained as the sum of the first order difference of the temporal codes. In this case, the count would be $1$, which indicated that this pig was scanned once (from deactivated zone into activated zone) and the total count should be added by $1$. Similarly, Fig.~\ref{fig:trajectory}d showed a pig trajectory with code [1, 0, 0, 0, 1, 1, 0, 0] and sum of the the first order difference inferred that the count was $-1$. This meant that this pig, which has been counted before, moved from scanned zone to to-be scanned zone. Thus, the total count should minus $1$. This design enabled the algorithm to avoid false positives counting caused by pig movements into/out camera view. Fig.~\ref{fig:trajectory}e-g showed examples when the pig trajectory count was $0$. Fig.~\ref{fig:trajectory}e-f represented pig trajectories that never went across the scanning line. Fig.~\ref{fig:trajectory}g represented cases where the trajectory started and ended in the same activity zone. These examples demonstrated that SFRT would not be influenced by the tracking failures (e.g. broke one trajectory into several cased by occlusion) that happened only in one single zone. In this study, a low-pass filter with window size of 5 was applied before the first order differential calculation. This low-pass filtering step was designed to avoid the trajectory jitter near the activity scanning line. The final counting result for the whole video also added the number of detected candidates in deactivated zone of the beginning frame and the number of detected candidates in activated zone of the ending frame.

\section{EXPERIMENTS}

% \begin{table}[htbp]
% 	\caption{Mean Absolute Percentage Error }
% 	\begin{center}
% 		\begin{tabular}{c|c|c|c|c|c|c|c}
% 			\hline
% 			%\textbf{Table}&\multicolumn{3}{|c|}{\textbf{Test}} \\
% 			%\cline{2-4} 
% 			\textbf{Architecture} & \textbf{U-Net}& \textbf{SegNet}& \textbf{Stacked Hourglass} & \textbf{Ours} & \textbf{Human*}  \\
% 			\hline
% 			\textbf{MAPE} & 2.59 & 2.93 & ??? & 3.11 & 11.0\\ 
% 			\hline
% 			\textbf{FLOPs} & ? & ? & ? & ? & N/A\\
% 			\hline			
% 			%\multicolumn{4}{l}{$^{\mathrm{a}}$ of a Table footnote.}
% 		\end{tabular}
% 		\label{tab:mape_nets}
% 	\end{center}
% \end{table}
	
% \subsection{DATASETS}
We collected 51 videos by inspection robots installed in pig grouping houses of two different pig farming corporations. All videos were originally recorded at 1280$\times$720 resolution with frame rate of 25 $f/s$. For this study, we first resized the video frame to 360$\times$640, and then cropped them to 352$\times$640. All experiments in this work used this resolution. Each video (pig house) had 120$\sim$250 pigs. The length of the videos ranged from 2 minutes to 4 minutes. We randomly split these videos into three subsets, 21 for training, 5 for validation and 25 for testing. The ground truth were provided by workers, who counted the pigs inside the grouping houses when the videos were recorded.

% We evaluate the keypoints detection accuracy using Percent of Detected Joints (PDJ) \cite{toshev2014deeppose}. A keypoint is considered as detected if the distance between the predicted keypoint and the ground truth as no longer than a fraction of the total length of the skeleton of the pig.
% 2) We evaluate the detection rate of the pig instances use Average Precision (AP) the compare different methods. A predicted bounding box is considered correct only if it has an intersection over union (IoU) no more than 0.5 with the ground truth bounding box. For any ground truth box, no more than one predicted box is considered as the correct prediction. All the surplus predicted boxes are treated as false alarms.
% 3) We report the error of pig counting using Mean Absolute Percentage Error (MAPE) and Mean Absolute Error (MAE).

\subsection{COMPARISON WITH HUMAN READER}

To demonstrate the effectiveness of out method, we compared the performance of our counting system with human readers on test dataset. There were three readers for this study. The readers were required to provide count results by watching the same top-down view videos as the input of the algorithm. There were no time limits for the reading process, and the readers were allowed to pause, rewind, replay the video and took notes for unlimited times. Each reader estimated the pig counts for all the videos in the test datasets. The counting error for both the proposed method and human reader were evaluated using mean absolute percentage error (MAPE) and mean absolute error (MAE). The three readers have MAPE of 11.0\%, 17.4\%, and 15.9\%; and MAE of 12.6, 26.3 and 25.2, respectively. The average time that the human readers have spent on per video is around 1.5 hours. In contrast, our method had MAPE of 2.67\%, and MAE of 3.32, which significantly outperformed the human readers.

% \subsection{Training details}
% Using the full resolution of the video frames is time and space consuming.
% We resize the images to a fixed width 640 pixels the height. At training stage, the images are randomly flipped horizontally and vertically. And then 352$\times$640 sub-images are cropped at random location on the fly to train the keypoint detection networks. We also apply color jittering to reduce overfitting. We use Adam~\cite{kingma2014adam} optimizer with no weight decay. The learning rate is initially set to ${10}^{-4}$ and decayed at the epoch 100 for each training. The batch size used is 24. The training from scratch is run for 120 epoches. Pretraining the encoder part of the proposed network (by adding an global average pooling layer followed by a fully connected layer) on Imagenet~\cite{deng2009imagenet} achieves similar results notwithstanding halving the training time. At testing stage, no augmentation is applied. A 352$\times$640 sub-image cropped from the center of each resized image to feed the networks.

\subsection{ABLATION STUDY}

To validate our proposed CNN architecture for keypoints detection, we compared our method with UNet~\cite{DBLP:conf/miccai/RonnebergerFB15} and stacked Hourglass network~\cite{newell2016stacked} using the same train, validation and test datasets. Both methods were modified to fit our pixel-level keypoints detection pipelines. Following ~\cite{isola2017image}, the cropping operators was removed from UNet and 7 UNet-submodules were used. The Stacked Hourglass network tested had two hourglass stacked. The Percent of Detected Joints (PDJ) \cite{toshev2014deeppose} was used as the evaluation metric. One keypoint was considered as detected if the distance between the predicted keypoint and the ground truth was smaller than a fraction of the total length of the skeleton of the pig. As  shown in Table ~\ref{tab:keypoint_performance}, our method achieved better keypoints detection accuracy for all 5 body parts with significantly less computation cost and smaller parameter size.

\begin{table}[t]
% \centering
\caption{Comparison of pig counting}
\begin{tabular}{l|l|l}
\hline
Method             & MAPE & MAE \\ \hline
% Human Testee       &  11.0  & 12.6\\ \hline
SSD \cite{liu2016ssd} + Tracking       &  327\%    &   412  \\
YOLOv3 \cite{redmon2018yolov3} + Tracking      &  247\%    &  368    \\
Proposed Keypoint Detector + Tracking      &   152\%   & 191 \\ \hline
SSD \cite{liu2016ssd} + Tracking + STRF  &  10.1\%    &   12.2  \\
YOLOv3 \cite{redmon2018yolov3} + Tracking + STRF &   5.35\%   &  7.00   \\
Proposed Keypoint Detector + Tracking + STRF &  \textbf{2.67}\%    &   \textbf{3.32}  \\ \hline % 3.29
\end{tabular}
\label{tab:counting_performance}
\end{table}

We also compared our bottom-up detection method with SSD~\cite{liu2016ssd} and YOLOv3 ~\cite{redmon2018yolov3} using top-down bounding boxes detection metric: mean average precision with 0.5 IOU (mAP@0.5). The proposed bottom-up approach did not directly output bounding boxes of pig. Thus, we used keypoints/skeleton bounding boxes instead. It should be noted that the keypoints bounding boxes are more strict and harder to predict, and our network was never trained for the bounding boxes detection task. SSD achieved 73.3\% mAP while YOLOv3 achieves 79.7\% mAP. Our method had 84.3\% mAP. Although more challenging, our method showed better performance. It should be noted that a large part of the detection failures happened around image boudaries where large fisheye distortion and image cutoff happened. Due to the design of STRF methods, most of the failures will not influence the final counting result.

% The proposed network for keypoint detection does not directly output bounding boxes. To compare SSD and YOLOv3, we generate boxes based on the keypoints and skeletons predicted by the proposed network using the following procedures: 1) Centered at each predicted keypoint we generate a circle of radius $r$. These circles are referred as "extended keypoint" 2) For each predicted pig instance, we draw the minimum bounding rectangle of the union of its extended keypoints as a bounding box. 3) For each predicted pig instance, the sum of the scores of all the detected keypoints is assigned to the bounding box. In our experiments, $r$ is simply set as 1/8 of the average length of pig skeletons across the training set and 84.3\% AP is achieved.

To evaluate the effectiveness of the STRF method, we compared the counting results with and without STRF using our detection method, SSD and YOLOv3, resepctively. Table \ref{tab:counting_performance} showed that the MAPE and MAE are significantly small when using STRF. And our method achieved better performance with/without STRF compared with SSD or YOLOv3.

% 3) We also try to localize pig instances using SSD and YOLOv3 instead of localizing the keypoints and skeletons. In this case we assign a temporal
% code to each predicted box based on its centroid. The results are shown in Table.~\ref{tab:counting_performance}.

% 4) \textbf{Directly using the total number of tracklets} as the predicted number of pigs we get a error rate 183\%. By checking the results frame by frame, we find that the overestimation occurs mainly because the tracking id reassignment to the same instances. When testing on the video clips each in length of 30 second which are manually selected from each video in test set and have relatively more crowding pigs and devices such as feeding machines, the error rate is as high as 249\%. 

\subsection{RUNTIME ANALYSIS}

We analyzed the runtime performance of our method using the test dataset. On desktop computer, it achieved 3.42 frames per second (FPS) running speed with a Intel i7-6850K CPU and 32GB DDR4 2133MHz Memory. When accelerated by a single NVIDIA GeForce GTX 1080Ti GPU, it achieved 82.6 FPS. The proposed counting algorithm has also been deployed on two different edge computing devices. It achieved 0.625 FPS on a Firefly-RK3399 platform, which had a 2GB Memory and a Rockchip RK3399 CPU. On NVIDIA Jetson Nano platform, it achieved 3.19 FPS with a 4GB memory, a quad-core ARM A57 CPU and 128 CUDA cores.

% %NVIDIA Titan RTX : 184.12 FPS regular FP32 mode. (Don't report because the GPU is so fast that the parts other than the CNN in the proposed system may become overhead.)
% 2) We also test the proposed system on Firefly-RK3399 platform, which has a 2GB Memory and a six-core Rockchip RK3399 CPU. On average takes 40 mins to process an 1 min video (1.60 seconds/Frame).

% 3) We also test the proposed system on NVIDIA Jetson Nano platform, which has a 4GB memory, a quad-core ARM A57 CPU and a 128 CUDA cores. The proposed system achieves 3.19 FPS.

\section{CONCLUSION}

In this work, we presented a hardware configuration and novel efficient algorithm for pig counting in large grouping houses. An inspection robot with a monocular fisheye camera was installed on the roof with rails, along which the root could move back and forth to collect top-down view videos. A novel efficient bottom-up CNN detection approach was developed to first detect pigs from the crowd. Second, a online tracking method was employed to associate pig ID temporally. A novel STRF method was proposed to calculate the final pig counts, while significantly avoid false positive counting due to tracking failure or large pig movements. The low computation cost design significantly reduce the computation time and model size. This counting algorithm has been deployed in edge computing device of the inspection robot, and achieved counting accuracy superior to human readers.

\addtolength{\textheight}{-12cm}   % This command serves to balance the column lengths

%%%%%%%%%%%%%%%%%%%%%%%%%%%%%%%%%%%%%%%%%%%%%%%%%%%%%%%%%%%%%%%%%%%%%%%%%%%%%%%%

%%%%%%%%%%%%%%%%%%%%%%%%%%%%%%%%%%%%%%%%%%%%%%%%%%%%%%%%%%%%%%%%%%%%%%%%%%%%%%%%



% \section*{ACKNOWLEDGMENT}

%%%%%%%%%%%%%%%%%%%%%%%%%%%%%%%%%%%%%%%%%%%%%%%%%%%%%%%%%%%%%%%%%%%%%%%%%%%%%%%%

\bibliographystyle{ieee}
\bibliography{references}

\begin{thebibliography}{10}\itemsep=-1pt

\bibitem{badrinarayanan2017segnet}
V.~Badrinarayanan, A.~Kendall, and R.~Cipolla.
\newblock Segnet: A deep convolutional encoder-decoder architecture for image
  segmentation.
\newblock {\em IEEE transactions on pattern analysis and machine intelligence},
  39(12):2481--2495, 2017.

\bibitem{cao2017realtime}
Z.~Cao, T.~Simon, S.-E. Wei, and Y.~Sheikh.
\newblock Realtime multi-person 2d pose estimation using part affinity fields.
\newblock In {\em Proceedings of the IEEE Conference on Computer Vision and
  Pattern Recognition}, pages 7291--7299, 2017.

\bibitem{gardenier2018object}
J.~Gardenier, J.~Underwood, and C.~Clark.
\newblock Object detection for cattle gait tracking.
\newblock In {\em 2018 IEEE International Conference on Robotics and Automation
  (ICRA)}, pages 2206--2213. IEEE, 2018.

\bibitem{hodgson2016precision}
J.~C. Hodgson, S.~M. Baylis, R.~Mott, A.~Herrod, and R.~H. Clarke.
\newblock Precision wildlife monitoring using unmanned aerial vehicles.
\newblock {\em Scientific reports}, 6:22574, 2016.

\bibitem{howard2017mobilenets}
A.~G. Howard, M.~Zhu, B.~Chen, D.~Kalenichenko, W.~Wang, T.~Weyand,
  M.~Andreetto, and H.~Adam.
\newblock Mobilenets: Efficient convolutional neural networks for mobile vision
  applications.
\newblock {\em arXiv preprint arXiv:1704.04861}, 2017.

\bibitem{isola2017image}
P.~Isola, J.-Y. Zhu, T.~Zhou, and A.~A. Efros.
\newblock Image-to-image translation with conditional adversarial networks.
\newblock In {\em IEEE Conference on Computer Vision and Pattern Recognition},
  2017.

\bibitem{liu2016ssd}
W.~Liu, D.~Anguelov, D.~Erhan, C.~Szegedy, S.~Reed, C.-Y. Fu, and A.~C. Berg.
\newblock Ssd: Single shot multibox detector.
\newblock In {\em European conference on computer vision}, pages 21--37.
  Springer, 2016.

\bibitem{liu2018robust}
X.~Liu, S.~W. Chen, S.~Aditya, N.~Sivakumar, S.~Dcunha, C.~Qu, C.~J. Taylor,
  J.~Das, and V.~Kumar.
\newblock Robust fruit counting: Combining deep learning, tracking, and
  structure from motion.
\newblock In {\em 2018 IEEE/RSJ International Conference on Intelligent Robots
  and Systems (IROS)}, pages 1045--1052. IEEE, 2018.

\bibitem{liu2019monocular}
X.~Liu, S.~W. Chen, C.~Liu, S.~S. Shivakumar, J.~Das, C.~J. Taylor,
  J.~Underwood, and V.~Kumar.
\newblock Monocular camera based fruit counting and mapping with semantic data
  association.
\newblock {\em IEEE Robotics and Automation Letters}, 4(3):2296--2303, 2019.

\bibitem{newell2016stacked}
A.~Newell, K.~Yang, and J.~Deng.
\newblock Stacked hourglass networks for human pose estimation.
\newblock In {\em European conference on computer vision}, pages 483--499.
  Springer, 2016.

\bibitem{papandreou2018personlab}
G.~Papandreou, T.~Zhu, L.-C. Chen, S.~Gidaris, J.~Tompson, and K.~Murphy.
\newblock Personlab: Person pose estimation and instance segmentation with a
  bottom-up, part-based, geometric embedding model.
\newblock In {\em Proceedings of the European Conference on Computer Vision
  (ECCV)}, pages 269--286, 2018.

\bibitem{DBLP:journals/cea/PezzuoloGSGM18}
A.~Pezzuolo, M.~Guarino, L.~Sartori, L.~A. Gonz{\'{a}}lez, and F.~Marinello.
\newblock On-barn pig weight estimation based on body measurements by a kinect
  v1 depth camera.
\newblock {\em Computers and Electronics in Agriculture}, 148:29--36, 2018.

\bibitem{redmon2018yolov3}
J.~Redmon and A.~Farhadi.
\newblock Yolov3: An incremental improvement.
\newblock {\em arXiv preprint arXiv:1804.02767}, 2018.

\bibitem{ren2015faster}
S.~Ren, K.~He, R.~Girshick, and J.~Sun.
\newblock Faster r-cnn: Towards real-time object detection with region proposal
  networks.
\newblock In {\em Advances in neural information processing systems}, pages
  91--99, 2015.

\bibitem{rivas2018detection}
A.~Rivas, P.~Chamoso, A.~Gonz{\'a}lez-Briones, and J.~Corchado.
\newblock Detection of cattle using drones and convolutional neural networks.
\newblock {\em Sensors}, 18(7):2048, 2018.

\bibitem{DBLP:conf/miccai/RonnebergerFB15}
O.~Ronneberger, P.~Fischer, and T.~Brox.
\newblock U-net: Convolutional networks for biomedical image segmentation.
\newblock In {\em MICCAI}, pages 234--241, 2015.

\bibitem{shen2018crowd}
Z.~Shen, Y.~Xu, B.~Ni, M.~Wang, J.~Hu, and X.~Yang.
\newblock Crowd counting via adversarial cross-scale consistency pursuit.
\newblock In {\em Proceedings of the IEEE conference on computer vision and
  pattern recognition}, pages 5245--5254, 2018.

\bibitem{sindagi2017generating}
V.~A. Sindagi and V.~M. Patel.
\newblock Generating high-quality crowd density maps using contextual pyramid
  cnns.
\newblock In {\em Proceedings of the IEEE International Conference on Computer
  Vision}, pages 1861--1870, 2017.

\bibitem{DBLP:journals/cea/TianGCWLM19}
M.~Tian, H.~Guo, H.~Chen, Q.~Wang, C.~Long, and Y.~Ma.
\newblock Automated pig counting using deep learning.
\newblock {\em Computers and Electronics in Agriculture}, 163, 2019.

\bibitem{toshev2014deeppose}
A.~Toshev and C.~Szegedy.
\newblock Deeppose: Human pose estimation via deep neural networks.
\newblock In {\em Proceedings of the IEEE conference on computer vision and
  pattern recognition}, pages 1653--1660, 2014.

\end{thebibliography}

\end{document}